\documentclass[sigconf,table,xcdraw]{acmart}
\AtBeginDocument{%
  }

\setcopyright{acmlicensed}
\copyrightyear{2025}
\acmYear{2025}
\acmDOI{XXXXXXX.XXXXXXX}
\acmConference[Conference acronym 'XX]{Make sure to enter the correct
  conference title from your rights confirmation email}{June 03--05,
  2018}{Woodstock, NY}
\acmISBN{978-1-4503-XXXX-X/2018/06}
\acmSubmissionID{3125}

\usepackage{booktabs}
\usepackage{graphicx}
\usepackage{subcaption}
\usepackage{multirow}

\usepackage{amsthm,amsmath,amsfonts,bm,xspace}
\usepackage{color}

\def\eg{{\em e.g.,}\xspace}
\def\ie{{\em i.e.,}\xspace}

\definecolor{mgreen}{rgb}{0,0.7,0}

\def\eqref#1{(\ref{#1})}

\def\1{\bm{1}}

\def\vp{{\bm{p}}}

\def\evp{{p}}

\def\mV{{\bm{V}}}

\DeclareMathAlphabet{\mathsfit}{\encodingdefault}{\sfdefault}{m}{sl}
\SetMathAlphabet{\mathsfit}{bold}{\encodingdefault}{\sfdefault}{bx}{n}

\def\sP{{\mathbb{P}}}

\newcommand{\system}[1]{\text{#1}}
\newcommand{\ourmethod}{\system{ReSR}\xspace}
\usepackage[normalem]{ulem}
\useunder{\uline}{\ul}{}

\begin{document}


\title{Physics-Grounded Motion Forecasting via Equation Discovery for Trajectory-Guided Image-to-Video Generation}


\author{Tao Feng}
\affiliation{%
  \institution{Monash University}
  \city{Melbourne}
  \country{Australia}}
\email{tao.feng@monash.edu}

\author{Xianbing Zhao}
\affiliation{%
  \institution{Harbin Institute of Technology}
  \city{Shezhen}
  \country{China}}
\email{zhaoxianbing_hitsz@163.com}

\author{Zhenhua Chen}
\affiliation{%
  \institution{Monash University}
  \city{Melbourne}
  \country{Australia}}
\email{zhenhua.chen@monash.edu}

\author{Tien Tsin Wong}
\affiliation{%
  \institution{Monash University}
  \city{Melbourne}
  \country{Australia}}
\email{TT.Wong@monash.edu}

\author{Hamid Rezatofighi}
\affiliation{%
  \institution{Monash University}
  \city{Melbourne}
  \country{Australia}}
\email{Hamid.Rezatofighi@monash.edu}

\author{Gholamreza Haffari}
\affiliation{%
  \institution{Monash University}
  \city{Melbourne}
  \country{Australia}}
\email{gholamreza.haffari@monash.edu}

\author{Lizhen Qu}
\authornote{Lizhen Qu is the corresponding author.}
\affiliation{%
  \institution{Monash University}
  \city{Melbourne}
  \country{Australia}}
\email{lizhen.qu@monash.edu}

\renewcommand{\shortauthors}{Trovato et al.}

\begin{abstract}
Recent advances in diffusion-based and autoregressive video generation models have achieved remarkable visual realism. However, these models typically lack accurate physical alignment, failing to replicate real-world dynamics in object motion. This limitation arises primarily from their reliance on learned statistical correlations rather than capturing mechanisms adhering to physical laws. To address this issue, we introduce a novel framework that integrates symbolic regression (SR) and trajectory-guided image-to-video (I2V) models for physics-grounded video forecasting. Our approach extracts motion trajectories from input videos, uses a retrieval-based pre-training mechanism to enhance symbolic regression, and discovers equations of motion to forecast physically accurate future trajectories. These trajectories then guide video generation without requiring fine-tuning of existing models. Evaluated on scenarios in Classical Mechanics, including spring-mass, pendulums, and projectile motions, our method successfully recovers ground-truth analytical equations and improves the physical alignment of generated videos over baseline methods.
\end{abstract}

\ccsdesc[500]{Computing methodologies~Scene understanding}


\keywords{Physics Learning, Equation Learning, Video Understanding, Video Generation}
\begin{teaserfigure}
  \includegraphics[width=\textwidth]{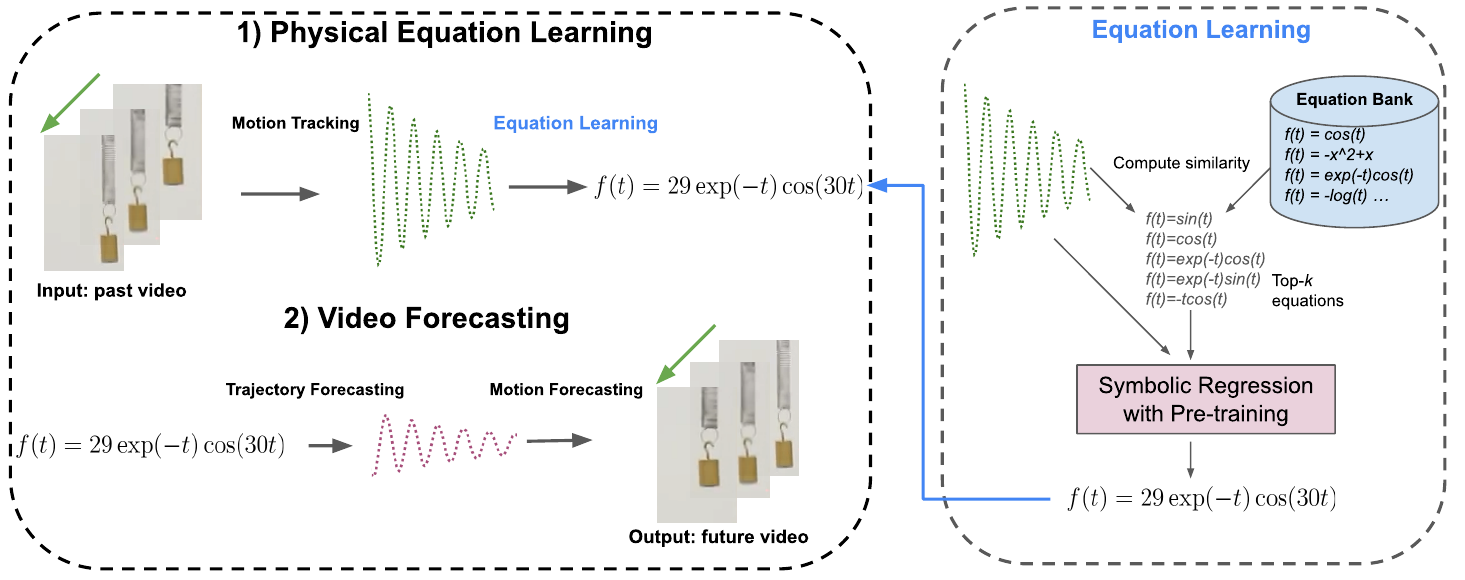}
  \caption{An overview of our proposed framework. Given an input video, we first extract object (\ie spring and weight) motion trajectories, which are used to discover governing equations of motion via symbolic regression enhanced by our proposed retrieval-based pre-training mechanism (\ourmethod). The learned symbolic equations forecast future object trajectories, serving as precise control signals to guide trajectory-guided video generation models, thus enabling physics-grounded video generation.}
    \label{fig:overall_pipeline}
\end{teaserfigure}


\maketitle


\section{Introduction}
Recent advances in video generation models have significantly improved the realism of synthesized videos, driven primarily by diffusion-based and autoregressive models \cite{blattmann2023stable, yang2025cogvideox, nvidia2025cosmosworldfoundationmodel, kong2025hunyuanvideosystematicframeworklarge}. Incorporating motion trajectories enables precise control over object movements, facilitating videos that more accurately capture intended dynamics~\cite{10.1007/978-3-031-72670-5_19, 10.1145/3641519.3657518, namekata2025sgiv}. However, existing trajectory-guided methods typically rely on text prompts, manually drawn or statistically derived trajectories~\cite{zhang2024toratrajectoryorienteddiffusiontransformer, klingai}, none of which ensures adherence to the underlying laws of physics~\cite{kang2024farvideogenerationworld, motamed2025generativevideomodelsunderstand, wang2025wisaworldsimulatorassistant}.

Physicists understand object dynamics by discovering physical laws from observational data and formulating these laws into symbolic equations. These equations reliably forecast object movements, unaffected by shifts in the underlying data distributions. Moreover, such equation discovery does not require extensive training data, unlike the scaling laws commonly adopted by current video generation models \cite{kaplan2020scalinglawsneurallanguage}. Therefore, for the \textit{first} time, we investigate: i) whether AI methods can feasibly discover physics equations directly from video clips and subsequently use these equations to reliably forecast object motion trajectories, and ii) whether such equations can be identified from just one or a handful of video clip without extensive data-driven training.

To address the above research questions, we propose a novel \textit{neuro-symbolic, inference-only} framework for forecasting object motion trajectories from a short video clip, followed by feeding the predicted trajectories into an image-to-video (I2V) model to produce physics-grounded videos. As illustrated in Figure~\ref{fig:overall_pipeline}, our approach first utilizes CoTracker~\cite{karaev2024cotracker3simplerbetterpoint} to extract initial object motion trajectories from a short video clip. We then employ a symbolic regression (SR) algorithm~\cite{cranmer2023interpretable}, an evolutionary search method that automatically discovers explicit mathematical equations, to derive a \textit{human-interpretable} symbolic equation characterizing the underlying physical law. Given the initial trajectories, this discovered equation can reliably produce future object movements of arbitrary length, consistently adhering to the underlying physics laws.

From another perspective, the equation discovery process can be viewed as training a symbolic model that characterizes motion trajectories. Current evolutionary search methods typically initialize their searches using randomly selected primitive functions, often starting far from the global optimum and resulting in slow convergence. To mitigate this, we propose a novel \textit{pre-training} method for symbolic regression, called \ourmethod, which initializes the search with a combination of the equations retrieved from an equation bank and randomly chosen functions. This approach substantially accelerates convergence and reduces generalization errors.


To investigate the fundamental challenges of learning equations from given video clip for the first time, we conduct experiments on a set of videos captured or synthesized in a controlled laboratory environment governed by the laws of Classical Mechanics. These videos depict systems, such as spring-mass oscillators, pendulums, and projectile motion. We choose this controlled setting because: i) it enables direct evaluation of discovered equations against ground-truth equations identified by physicists; ii) insights into object motion in Classical Mechanics can be easily extended or integrated into other types of motion; and iii) Classical Mechanics underpins a wide range of real-world applications, including physics simulation, scientific visualization, and physics education.

Our contributions are summarized as follows:
\begin{itemize}
    \item We propose a novel neuro-symbolic framework for physics-grounded motion forecasting, which combines SR with trajectory-guided video synthesis. Our approach operates entirely at inference time and does not require fine-tuning of video generation models.
    \item We introduce a \textit{retrieval-based pre-training mechanism} for SR, denoted as \ourmethod, which significantly improves convergence speed and accuracy in discovering equations from observed trajectories.
    \item We demonstrate empirically that \ourmethod successfully recovers equations that closely align with both ground-truth analytical expressions and empirical trajectories. Furthermore, videos generated using our framework exhibit stronger physical consistency compared to existing baselines. 
\end{itemize}

\section{Background Knowledge}
\paragraph{Symbolic Regression.}
Throughout science history, scientists have discovered empirical laws from observational data. For example, Johannes Kepler formulated the third law of planetary motion, $(\text{period})^2\propto (\text{radius})^3$, after analyzing thirty years of astronomical data. Similarly, Planck’s law was a function fitted to experimental data \cite{planck1900ueber}. However, modern scientific data is often high-dimensional and complex, making manual equation discovery a challenging task \cite{virgolin2022symbolic}. 

SR has emerged as a computational approach to automatically discover mathematical equations from obercational data. Unlike traditional regression, which fits data to a predefined equation structure (\eg linear or polynomial regression), SR searches for both the equation structure and corresponding parameters. This flexibility makes SR particularly valuable in scientific discovery, where both interpretability and accuracy are crucial \cite{doi:10.1126/sciadv.1602614, meidani2023identification}. 

There are several approaches to SR, with two primary categories: evolutionary algorithm (EA)-based methods and deep learning-based methods. EA-based approaches are more common and operate by evolving a population of candidate equations as tree structures over successive generations, using operations such as mutation and crossover to search for equations that best fit the data \cite{brindle1980genetic, goldberg1991comparative, doi:10.1137/0218082, gplearn, cranmer2023interpretablemachinelearningscience}. The typical workflow of EA-based SR follows these steps: 1) Initialization: generate an initial population of random equations. 2) Selection: randomly sample a subset of equations from the population. 3) Parameter Optimization: optimize numerical parameters within each equation. 4) Evaluation: compute a fitness score for each equation by comparing its predictions to the observed data. 5) Mutation \& Crossover: modify equations through mutation (randomly altering parts of the equation) and crossover (combining parts of different equations). 6) Replacement: the least-fit equations are replaced with new ones, and the process repeats until convergence. One major advantage of EA-based methods is their ability to run efficiently on multi-core CPUs, making them more accessible and scalable for large datasets. In contrast, deep learning-based symbolic regression methods typically require GPU acceleration, which can be significantly more time- and resource-intensive. Additionally, EA-based methods require minimal prior assumptions about the functional form of the equation, allowing them to explore a diverse solution space.

Deep learning-based methods directly predict equations from data \cite{biggio2021neural, kamienny2022endtoend, shojaee2023transformerbased, meidani2024snip}. Inspired by the success of transformer-based models \cite{devlin-etal-2019-bert, Radford2019LanguageMA, feng-etal-2023-less, feng-etal-2025-causalscore}, deep learning-based methods typically train an end-to-end transformer-based model where the input is observational data and the output is a symbolic equation. However, such methods face several limitations. First, deep learning models often struggle with out-of-distribution generalization \cite{yang2024generalized, kim2024investigation, feng-etal-2024-imo}. Real-world scientific data is usually novel and noisy, out of the scope of training distributions, which makes models difficult to generate accurate equations for real scientific data. Second, there is no guarantee that the generated output forms a syntactically valid equation, potentially leading to non-executable equations. 

Some approaches attempt to combine deep learning and EA \cite{landajuela2022a, 8469901, grayeli2024symbolic}. These methods use deep learning models to generate equation candidates or guide evolutionary search. However, such methods often require a high computational cost.  Deep learning models must first be trained on large-scale datasets and then utilized during each iteration of the evolutionary process, resulting in significant resource demands. In contrast, pure EA-based methods can run efficiently on CPU machines.

Inspired by the success of pre-training in deep learning \cite{pmlr-v9-erhan10a, devlin-etal-2019-bert}, we propose a retrieval-based pre-training mechanism for EA-based SR (see Section~\ref{sec:method_symbolic_regression}). We first construct an equation bank containing physics-related equations. During initialization, the SR algorithm retrieves equations that closely align with the observed data and uses them as initial candidates. This pre-training strategy significantly improves convergence speed and enhances the physical alignment of the learned equations.

\paragraph{Trajectory-Guided Video Generation.}
Recent advances in video generation have led to models capable of producing visually compelling videos \cite{yan2021videogptvideogenerationusing, 9878449, 10377456, shen2023storygptvlargelanguagemodels, wang2024emu3nexttokenpredictionneed, yang2025cogvideox}. However, recent studies \cite{kang2024farvideogenerationworld, motamed2025generativevideomodelsunderstand, wang2025wisaworldsimulatorassistant} have shown that current text-to-video (T2V) and I2V models often fail to generate videos that align with real-world physics, particularly in representing accurate object motion. 

Trajectory-guided video generation is a motion-aware video synthesis framework in which object movement is explicitly controlled by numerical trajectories, which are typically represented as sequences of $(x,y)$ coordinates over time. Most existing trajectory-guided video generation models are built on diffusion-based architectures \cite{xing2025motioncanvas,10.5555/3495724.3496298, song2020score,  10.1145/3641519.3657518, 10.1007/978-3-031-72670-5_19, namekata2025sgiv, fu2025dtrajmaster, zhang2024toratrajectoryorienteddiffusiontransformer}, where the input consists of an initial image as the first frame and trajectories indicating object motion. However, in prior work, these trajectories are manually drawn, which does not ensure alignment with real-world dynamics. In contrast, we use learned equations from observational data to generate future trajectories, ensuring that the future object motion follows discovered physical dynamics. By conditioning on these physics-aligned trajectories, we aim to use trajectory-guided video generation models to predict future videos that align with real physical dynamics.

\section{Methodology}

\subsection{Task Formulation and Notations}

The objective of this study is to achieve physics-grounded motion forecasting for trajectory-guided video generation. As illustrated in Figure~\ref{fig:overall_pipeline}, given an input video $\mV_{i}$ depicting the initial motion of an object, our approach generates a video $\mV_{o}$ representing the object's future motion. Our approach consists of three main steps. First, we extract the motion trajectories of moving objects in $\mV_{i}$. The extracted trajectories are represented as a set $\sP = \{\vp_1, \vp_2, ..., \vp_n \}$, where each trajectory $\vp_i$ is a time series of object positions: $\vp_i = \left [ \evp_1, \evp_2, ..., \evp_T \right ]$, where $\evp_t = (x_t, y_t)$ denotes the image-space coordinate of the object at time step $t$. Next, we employ symbolic regression to learn equations that govern the motion of objects. Specifically, for each trajectory $\vp_i$, we aim to learn a pair of functions $f_i^x(t)$ and $f_i^y(t)$ such that:
\begin{equation}
\label{equ:mapping_function}
x_t = f_i^x(t), \quad y_t = f_i^y(t).    
\end{equation}
that map time to object position. Using the learned equation $f_i^x(t)$ and $f_i^y(t)$, we predict the future trajectory for time steps beyond the observed interval, i.e., $\vp_i=f_{i}(t), t\in \{ T+1, T+2, ..., T+K \}$, where $K$ represents the forecast horizon. Finally, we utilize the predicted trajectories to guide trajectory-based video generation models, which then synthesize the future video $\mV_{o}$.

\subsection{Extraction of Object Motion Trajectory}
To learn equations of object motion, we first extract object trajectories from the input video $\mV_{i}$. We employ CoTracker \cite{karaev2024cotracker3simplerbetterpoint}, a state-of-the-art point tracking model that performs joint point tracking and propagation across all frames.  CoTracker requires a set of query points in the first frame to initiate tracking. While manual annotation is possible, it is not scalable across diverse video content. Instead, we adopt an automated approach by uniformly sampling query points on a 2D $M \times M$ grid across the first frame. Each query point is tracked throughout the entire video. We perform all tracking in the original image coordinate system without additional preprocessing. After collecting all trajectories, we compute the temporal variance of each trajectory. We then rank the trajectories based on their positional variance across time and retain the top-$k$ trajectories with the highest motion magnitude. This strategy is motivated by the observation that target objects in physics-driven videos typically exhibit the most motion, while background regions tend to remain static. As a result, selecting high-variance trajectories increases the likelihood of capturing the true object dynamics and filtering out irrelevant background noise.



\subsection{Symbolic Regression with Pre-training}
\label{sec:method_symbolic_regression}

In this step, we apply symbolic regression with retrieval-based pre-training (\ourmethod) to discover equations that fit extracted object trajectories. Instead of initializing the search process from scratch with random equations, we retrieve a set of candidate equations from a curated equation bank composed of physics-related equations. The retrieved equations then serve as priors to initialize the symbolic regression.  Given a trajectory $\vp_i = [\evp_1, \evp_2, ..., \evp_T]$, our goal is to learn Equation~\ref{equ:mapping_function}. 



\paragraph{Construction of Equation Bank}

We construct an equation bank containing a diverse set of equations derived from classical and empirical physics to serve as priors for motion dynamics. The bank integrates equations from three main sources: 
1) The Feynman equation dataset \cite{udrescu2020ai}, which consists of equations extracted from the Feynman Lectures on Physics \cite{feynman1965flp}. These equations typically take the form $y = f(x_1, x_2, \dots)$, with up to ten input variables. To adapt them for time-series motion, we substitute time-dependent variables (\eg velocity, acceleration, momentum) with time variable $t$. Variables that are independent of time (\eg mass, density) are replaced with constant values (\eg 10), aiming to preserve functional structure and remove irrelevant variables.
2) The Nguyen dataset \cite{uy2011semantically}, which includes commonly used empirical formulas in symbolic regression benchmarks. We apply the same time-variable substitution process.
3) A set of manually augmented physics equations not included in the above datasets. These cover well-known motion dynamics such as (damped) harmonic oscillators and projectile motions, ensuring the equation bank includes representative equations for various physical systems.

All equations are stored as symbolic expressions in Julia syntax \cite{bezanson2017julia}, enabling compatibility with our symbolic regression framework. Table~\ref{tab:statistics_equation_bank} summarizes the composition of the equation bank.

\begin{table}[ht]
\centering
\resizebox{\columnwidth}{!}{%
\begin{tabular}{lccc}
\hline
\textbf{Source} & \textbf{\# Equations} & \textbf{Avg, Equation Length} & \textbf{Preprocessing} \\
Feynman         & 106                  & 19.7                          & variable conversion    \\
Nguyen          & 10                   & 11.2                          & variable conversion    \\
Augmented       & 13                   & 10.3                          & N/A                    \\ \hline
\end{tabular}%
}
\caption{Statistics of equation bank.}
\label{tab:statistics_equation_bank}
\vspace{-15pt}
\end{table}

\paragraph{Retrieval-based Pre-training Mechanism}

Our proposed \ourmethod initializes symbolic regression with candidate equations retrieved from a curated equation bank. The retrieval is based on the similarity between the extracted object trajectory and trajectories generated by each equation in the bank. Similarity is computed using Dynamic Time Warping (DTW) \cite{muller2007dynamic}, a sequence alignment algorithm that handles temporal misalignments such as phase shifts and local time warping that are not captured by Euclidean distance.

However, standard DTW is unable to robustly handle spatial offsets and scale variations in trajectory coordinates. To address this, we introduce \textit{Normalized Dynamic Time Warping} (N-DTW), where the extracted trajectory is rescaled to match the coordinate range of each equation-generated trajectory before computing DTW. This helps the comparison to focus on shape similarity rather than absolute position. Formally, given an extracted trajectory \( \vp_i = [\evp_1, \evp_2, ..., \evp_T] \), where each \( \evp_t = (x_t, y_t) \), we normalize it as follows:
\begin{align}
\bar{x}_t &= (\hat{x}_{\max} - \hat{x}_{\min}) \cdot \frac{x_t - x_{\min}}{x_{\max} - x_{\min}} + \hat{x}_{\min} \\
\bar{y}_t &= (\hat{y}_{\max} - \hat{y}_{\min}) \cdot \frac{y_t - y_{\min}}{y_{\max} - y_{\min}} + \hat{y}_{\min}
\end{align}
where \( x_{\min}, x_{\max}, y_{\min}, y_{\max} \) are the bounds of the extracted trajectory, and \( \hat{x}_{\min}, \hat{x}_{\max}, \hat{y}_{\min}, \hat{y}_{\max} \) are the bounds of the equation-generated trajectory.

For each equation in the bank, we compute an N-DTW score with the normalized extracted trajectory. Since the similarity between the extracted trajectory and each equation-generated trajectory is computed independently, N-DTW retrieval can be easily parallelized across multiple CPU cores, enabling scalability to large equation banks. We then select the top-\textit{k} equations with the lowest distances as initial candidates for symbolic regression. This retrieval strategy emphasizes shape similarity rather than proximity in raw values. For instance, consider a trajectory generated by \( y = 0.5\cos(t + 3) + 100 \). Two candidate equations might be \( y = 100 \) and \( y = \cos(t) \). While Euclidean distance may favor \( y = 100 \) due to its proximity in magnitude, it fails to capture the oscillatory structure. In contrast, N-DTW correctly identifies \( y = \cos(t) \) as the more structurally similar trajectory.


\paragraph{Initialization of \ourmethod}
We initialize a portion of population members with the top-\textit{k} retrieved equations that closely match the target trajectory. We introduce an initialization weight hyperparameter \( \alpha \in [0, 1] \), which determines the proportion of initial population members that are seeded with retrieved equations, while the remaining are randomly generated. This hybrid initialization strategy allows us to balance \textit{exploration}—via randomly sampled equations that enable diversity in the search space—and \textit{exploitation}—via retrieved equations that act as informative priors. Higher values of \( \alpha \) prioritize faster convergence, while lower values preserve the capacity for discovering novel equation forms. If the available number of top-\textit{k} retrieved equations is insufficient to meet the required number based on $\alpha$, we duplicate top-\textit{k} retrieved equations to fill the remaining positions. This strategy ensures that the initial population predominantly contains equations closely matching the observed dynamics, reducing the risk of including irrelevant or misleading equations that could negatively impact the search efficiency.
This initialization occurs only once at the beginning of the symbolic regression run. All modifications, including retrieval-based pre-training and the integration of N-DTW, are implemented within a modified version of the \texttt{SymbolicRegression.jl} framework \cite{cranmer2023interpretablemachinelearningscience}, ensuring compatibility with existing symbolic regression workflows and reproducibility of our method.

\subsection{Trajectory-Guided Video Forecasting}
To generate future video frames $\mV_{o}$ that are physically consistent with learned motion dynamics, we incorporate existing trajectory-guided I2V models, such as SG-I2V \cite{namekata2025sgiv}, Tora \cite{zhang2024toratrajectoryorienteddiffusiontransformer}, and MotionCtrl \cite{10.1145/3641519.3657518}, into our framework. These models are typically diffusion models \cite{song2020score} that synthesize temporally coherent video sequences by denoising noise-perturbed images conditioned on a starting image and motion trajectories.

We use the final frame of the observed input video $\mV_{i}$ as the starting image and condition on future trajectories predicted by equations learned from \ourmethod. These trajectories are formatted as sequences of \( (x, y) \) coordinates, sampled at temporal intervals that match the requirement of each I2V model. This integration enables our framework to produce future video sequences that are not only visually plausible but also governed by equations of motion inferred from past observations. Our approach is \textit{modular} and \textit{model-agnostic}: it can be directly applied to any trajectory-guided I2V model without retraining or fine-tuning.

\section{Experiments}
We conduct two sets of experiments to evaluate the effectiveness of our framework. First, we assess whether the proposed \ourmethod enhances the performance of symbolic regression in discovering accurate motion equations. Second, we evaluate motion forecasting in the context of trajectory-guided video generation to determine whether videos generated from trajectories predicted by the learned equations exhibit improved alignment with real-world physical dynamics. For video capture in the controlled laboratory environment, we assume: 1) object motion is restricted to a 2D plane and recorded from an orthographic viewpoint, and 2) object trajectories are fully observable throughout the entire video duration.

\subsection{Evaluation of Symbolic Regression}
\label{sec:eval_SR}
\paragraph{Datasets.}
We evaluate symbolic regression performance using trajectories extracted from videos of classical physics systems, divided into two categories: \textbf{1)} \textit{systems with ground-truth trajectory equations}  (\ie systems with analytical solutions), including spring mass, damped spring mass, two body, and projectile motion \cite{huang2024automateddiscoverycontinuousdynamics}; \textbf{2)} \textit{systems without ground-truth trajectory equations}, including single pendulum, double pendulum and fluid motion, where no closed-form analytical solution exists \cite{huang2024automateddiscoverycontinuousdynamics, ohana2024the}. Each system includes ten videos with varying initial states. We use CoTracker \cite{karaev2024cotracker3simplerbetterpoint} to extract uniformly sampling query points on a $10 \times 10$ grid from the first frame. From these, we select the top 5 trajectories with the highest temporal variance to serve as inputs for symbolic regression methods, which aim to learn the equations of these trajectories. Each trajectory is split 80\%/10\%/10\% along the time dimension: the first 80\% for equation learning, the middle 10\% for validation, and the final 10\% for evaluation. Since trajectories are time-series data, this split aims to select equations that generalize from past states to unseen future states.

\paragraph{Evaluation of Learned Equations.}
For systems with ground-truth equations, we evaluate symbolic similarity between predicted equations and ground-truth equations using normalized Tree Edit Distance (TED) \cite{doi:10.1137/0218082}, which measures how many edit operations (\ie insertions, deletions, substitutions) are required to transform one equation tree into another, normalized by the maximum node count of two equation trees. For systems without ground-truth equations, we measure the Mean Squared Error (MSE) between the trajectory generated by predicted equations and the actual observed trajectory. Additionally, to compare the convergence of different symbolic regression methods, we record the MSE at each iteration.

\paragraph{Baselines}
We compare against the following methods: 
\textbf{APO} \cite{10.1145/1830483.1830584}: A symbolic regression method using Age-fitness Pareto Optimization. 
\textbf{gplearn} \cite{gplearn}: An EA-based symbolic regression with a scikit-learn-style API. 
\textbf{uDSR} \cite{landajuela2022a}: A hybrid approach that combines deep learning models with evolutionary algorithms to discover equations. 
\textbf{KAN} \cite{liu2025kan}: Kolmogorov-Arnold Networks (KANs) replace each weight in Multi-Layer Perceptrons (MLPs) with a univariate function parameterized as a spline, enabling symbolic equation extraction after training. 
\textbf{PySR} \cite{cranmer2023interpretablemachinelearningscience}: A symbolic regression framework based on evolutionary search, which can be viewed as an ablation model without retrieval-based pre-training. 
\textbf{LaSR} \cite{grayeli2024symbolic}: A symbolic regression approach that leverages large language models to propose initial equations.
We also perform an ablation study on \ourmethod using various initialization weights $\alpha$.

\paragraph{Implementation Details}
For symbolic regression methods using evolutionary algorithms, including both our method and baselines, we run 100 iterations with a population size of 30 across 30 populations. The search space operators include basic arithmetic (+, -, *, /), power functions, and common mathematical functions: cos, sin, exp, log, tan, and sqrt. For baselines that do not use evolutionary algorithms, such as KAN, we perform grid-based hyperparameter tuning and report results using the best-performing configuration on the validation set. All experiments are conducted on a machine with 32-core CPUs and a single 80GB A100 GPU.

\paragraph{Results and Analysis}

Table~\ref{tab:eval_SR} reports the performance of symbolic regression methods across classical physics systems. \ourmethod consistently outperforms all baselines in both symbolic similarity (TED) and trajectory error (MSE), demonstrating improved accuracy in discovering physical equations. Notably, \ourmethod achieves the best results across all systems, including those without analytical solutions (\eg single/double pendulum and fluid dynamics), where equation discovery is more challenging.

For the ablation study, we analyze the effect of the initialization weight hyperparameter $\alpha$, which controls the proportion of retrieved equations used to initialize the population. As shown in Table~\ref{tab:eval_SR}, performance improves steadily as $\alpha$ increases, peaking at $\alpha=0.75$, where 75\% of the population is seeded with retrieved equations and the rest randomly initialized. This balance supports both exploitation (using physics-aligned priors) and exploration (diversity through random sampling). Interestingly, setting $\alpha=1.0$ slightly degrades performance on some systems, likely due to over-reliance on retrieved equations that may not fully cover the solution space, especially for systems without analytical solutions.

\begin{table}[ht]
\centering
\resizebox{\linewidth}{!}{%
\begin{tabular}{@{}lccccccc@{}}
\toprule
                                   & \multicolumn{4}{c|}{\textbf{Systems with GT Equations}}                                                                                                                                                                       & \multicolumn{3}{c}{\textbf{Systems w/o GT Equations}}                                                                                   \\
\multirow{-2}{*}{\textbf{Methods}} & SM                                                                & DampSM                                     & TwoB                                       & \multicolumn{1}{c|}{Proje.}                                     & SP                                          & DP                                          & Fluid                                       \\ \midrule
\multicolumn{8}{c}{\textbf{Symbolic Equation Similarity (TED) (↑)}}                                                                                                                                                                                                                                                                                                                                          \\ \midrule
APO                                & $0.35_{0.11}$                                               & $0.25_{0.10}$                         & $0.35_{0.11}$                         & \multicolumn{1}{c|}{$0.35_{0.12}$}                         & N/A                                         & N/A                                         & N/A                                         \\
gplearn                            & $0.41_{ 0.11}$                                               & $0.35_{0.09}$                         & $0.46_{0.08}$                         & \multicolumn{1}{c|}{$0.39_{0.12}$}                         & N/A                                         & N/A                                         & N/A                                         \\
uDSR                               & $0.42_{ 0.12}$                                               & $0.32_{0.14}$                         & $0.45_{0.12}$                         & \multicolumn{1}{c|}{$0.42_{0.11}$}                         & N/A                                         & N/A                                         & N/A                                         \\
KAN                                & $0.31_{ 0.10}$                                               & $0.13_{0.14}$                         & $0.23_{0.14}$                         & \multicolumn{1}{c|}{$0.21_{0.17}$}                         & N/A                                         & N/A                                         & N/A                                         \\
PySR                               & $0.55_{ 0.16}$                                               & $0.38_{0.13}$                         & $0.51_{0.18}$                         & \multicolumn{1}{c|}{$0.43_{0.15}$}                         & N/A                                         & N/A                                         & N/A                                         \\
LaSR                               & $0.61_{0.15}$                                                & $0.43_{0.14}$                         & $0.56_{0.16}$                         & \multicolumn{1}{c|}{$0.56_{0.14}$}                         & N/A                                         & N/A                                         & N/A                                         \\
\textbf{\ourmethod-$0$}            & $0.55_{ 0.16}$                                               & $0.38_{0.13}$                         & $0.51_{0.18}$                         & \multicolumn{1}{c|}{$0.43_{0.15}$}                         & N/A                                         & N/A                                         & N/A                                         \\
\textbf{\ourmethod-$0.25$}         & $0.68_{0.12}$                                                & $0.51_{0.17}$                         & $0.59_{0.11}$                         & \multicolumn{1}{c|}{$0.61_{0.11}$}                         & N/A                                         & N/A                                         & N/A                                         \\
\textbf{\ourmethod-$0.5$}          & $0.74_{0.12}$                                                & $0.64_{0.11}$                         & $0.71_{0.14}$                         & \multicolumn{1}{c|}{$0.68_{0.12}$}                         & N/A                                         & N/A                                         & N/A                                         \\
\textbf{\ourmethod-$0.75$}         & \cellcolor[HTML]{38FFF8}{\color[HTML]{333333} $0.87_{0.08}$} & \cellcolor[HTML]{38FFF8}$0.76_{0.09}$ & \cellcolor[HTML]{FFFFC7}$0.79_{0.12}$ & \multicolumn{1}{c|}{\cellcolor[HTML]{38FFF8}$0.79_{0.05}$} & N/A                                         & N/A                                         & N/A                                         \\
\textbf{\ourmethod-$1.0$}          & \cellcolor[HTML]{FFFFC7}$0.81_{0.09}$                        & \cellcolor[HTML]{FFFFC7}$0.70_{0.10}$ & \cellcolor[HTML]{38FFF8}$0.83_{0.11}$ & \multicolumn{1}{c|}{\cellcolor[HTML]{FFFFC7}$0.73_{0.08}$} & N/A                                         & N/A                                         & N/A                                         \\ \midrule
\multicolumn{8}{c}{\textbf{Trajectory Error (MSE) (↓)}}                                                                                                                                                                                                                                                                                                                                                      \\ \midrule
APO                                & $6.76_{0.23}$                                                & $7.39_{0.14}$                         & $8.20_{0.28}$                         & \multicolumn{1}{c|}{$9.34_{0.11}$}                         & $73.35_{9.46}$                         & $75.29_{4.29}$                         & $80.93_{8.92}$                         \\
gplearn                            & $3.74_{0.13}$                                                & $3.59_{0.24}$                         & $4.21_{0.21}$                         & \multicolumn{1}{c|}{$3.93_{0.26}$}                         & $49.39_{8.20}$                         & $54.27_{9.17}$                         & $82.18_{4.94}$                         \\
uDSR                               & $3.11_{0.06}$                                                & $3.76_{0.11}$                         & $3.93_{0.13}$                         & \multicolumn{1}{c|}{$4.12_{0.15}$}                         & $39.58_{7.15}$                         & $45.07_{5.90}$                         & $66.39_{5.24}$                         \\
KAN                                & $10.34_{0.22}$                                               & $9.99_{0.51}$                         & $11.53_{0.43}$                        & \multicolumn{1}{c|}{$12.70_{0.81}$}                        & $85.42_{9.35}$                         & $93.72_{9.43}$                         & $95.14_{9.93}$                         \\
PySR                               & $2.73_{0.03}$                                                & $2.93_{0.07}$                         & $3.12_{0.05}$                         & \multicolumn{1}{c|}{$3.01_{0.03}$}                         & $26.13_{3.64}$                         & $33.55_{4.43}$                         & $76.10_{5.11}$                         \\
LaSR                               & $1.81_{0.07}$                                                & $1.95_{0.04}$                         & $1.91_{0.03}$                         & \multicolumn{1}{c|}{$1.93_{0.04}$}                         & $15.93_{3.30}$                         & $21.50_{2.36}$                         & $60.26_{6.49}$                         \\
\textbf{\ourmethod-$0$}            & $2.73_{0.03}$                                                & $2.93_{0.07}$                         & $3.12_{0.05}$                         & \multicolumn{1}{c|}{$3.01_{0.03}$}                         & $26.13_{3.64}$                         & $33.55_{4.43}$                         & $76.10_{5.11}$                         \\
\textbf{\ourmethod-$0.25$}         & $1.72_{0.08}$                                                & $1.91_{0.07}$                         & $1.85_{0.04}$                         & \multicolumn{1}{c|}{$1.87_{0.06}$}                         & $14.06_{2.13}$                         & $20.15_{2.25}$                         & $58.49_{5.10}$                         \\
\textbf{\ourmethod-$0.5$}          & $1.61_{0.05}$                                                & $1.84_{0.04}$                         & $1.68_{0.06}$                         & \multicolumn{1}{c|}{$1.83_{0.05}$}                         & $12.15_{2.36}$                         & $18.15_{3.21}$                         & \cellcolor[HTML]{38FFF8}$54.26_{3.78}$ \\
\textbf{\ourmethod-$0.75$}         & \cellcolor[HTML]{38FFF8}$1.36_{0.03}$                        & \cellcolor[HTML]{38FFF8}$1.55_{0.04}$ & \cellcolor[HTML]{FFFFC7}$1.46_{0.06}$ & \multicolumn{1}{c|}{\cellcolor[HTML]{38FFF8}$1.70_{0.03}$} & \cellcolor[HTML]{38FFF8}$10.46_{2.12}$ & \cellcolor[HTML]{38FFF8}$16.13_{3.41}$ & $56.14_{5.31}$                         \\
\textbf{\ourmethod-$1.0$}          & \cellcolor[HTML]{FFFFC7}$1.46_{0.02}$                        & \cellcolor[HTML]{FFFFC7}$1.61_{0.05}$ & \cellcolor[HTML]{38FFF8}$1.37_{0.03}$ & \multicolumn{1}{c|}{\cellcolor[HTML]{FFFFC7}$1.77_{0.04}$} & \cellcolor[HTML]{FFFFC7}$11.93_{2.42}$ & \cellcolor[HTML]{FFFFC7}$17.83_{2.15}$ & \cellcolor[HTML]{FFFFC7}$55.21_{4.16}$ \\ \bottomrule
\end{tabular}%
}
\caption{Evaluation of symbolic regression methods. The number following \ourmethod denotes the initialization weight hyperparameter, $\alpha$. GT: ground-truth, SM: Spring-Mass, TwoB: Two-Body, Proje: Projectile, SP: Single Pendulum, DP: Double Pendulum. The top two results are highlighted as \colorbox[HTML]{38FFF8}{best} and \colorbox[HTML]{FFFFC7}{second}, respectively. }
\label{tab:eval_SR}
\vspace{-15pt}
\end{table}

\begin{figure}[ht]
    \centering
    \begin{subfigure}[a]{\columnwidth}
        \centering
        \includegraphics[width=\columnwidth]{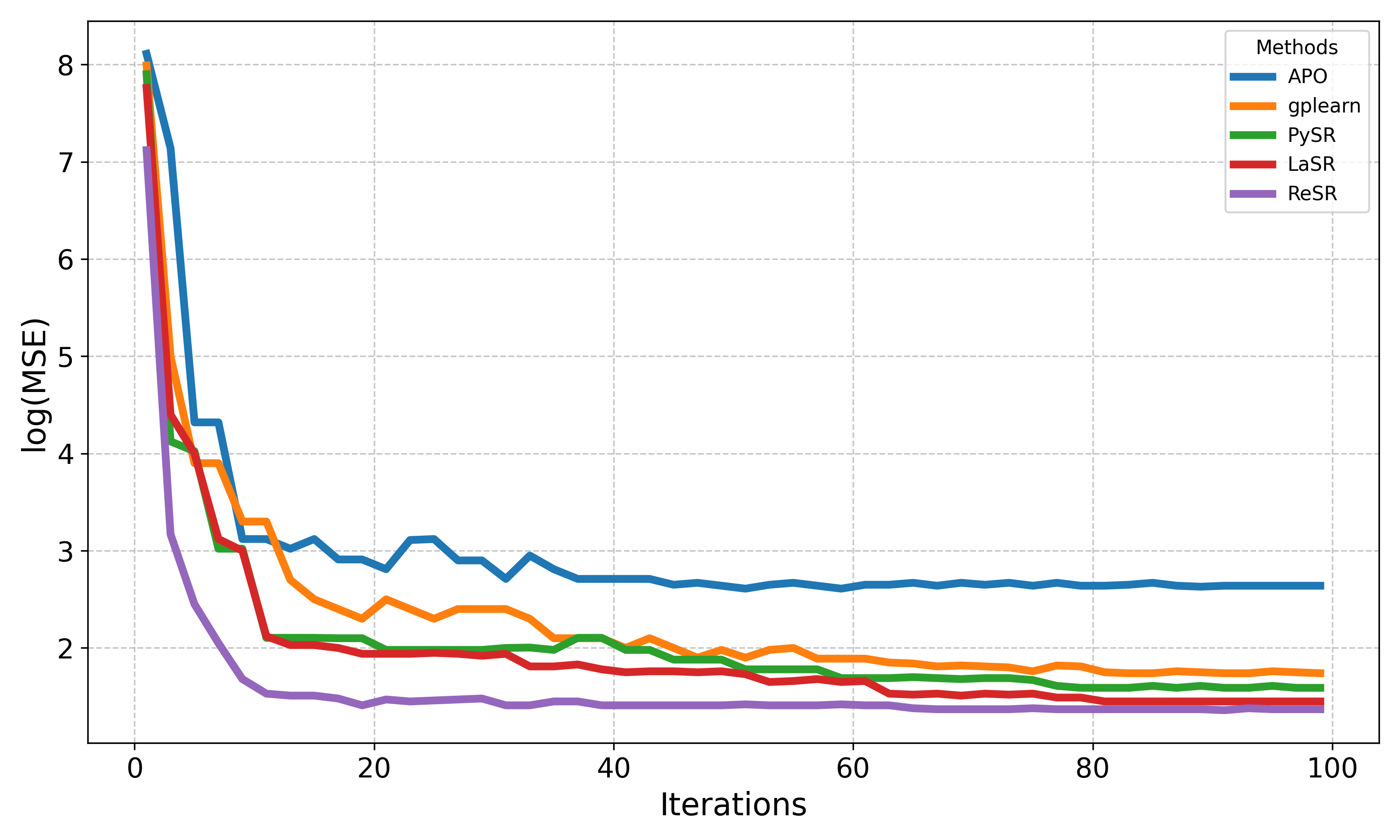}
        \caption{Training error (MSE) over iterations.}
        \label{fig:SR_convergence_train}
    \end{subfigure}
    
    \vspace{0.1cm}  
    
    \begin{subfigure}[b]{\columnwidth}
        \centering
        \includegraphics[width=\columnwidth]{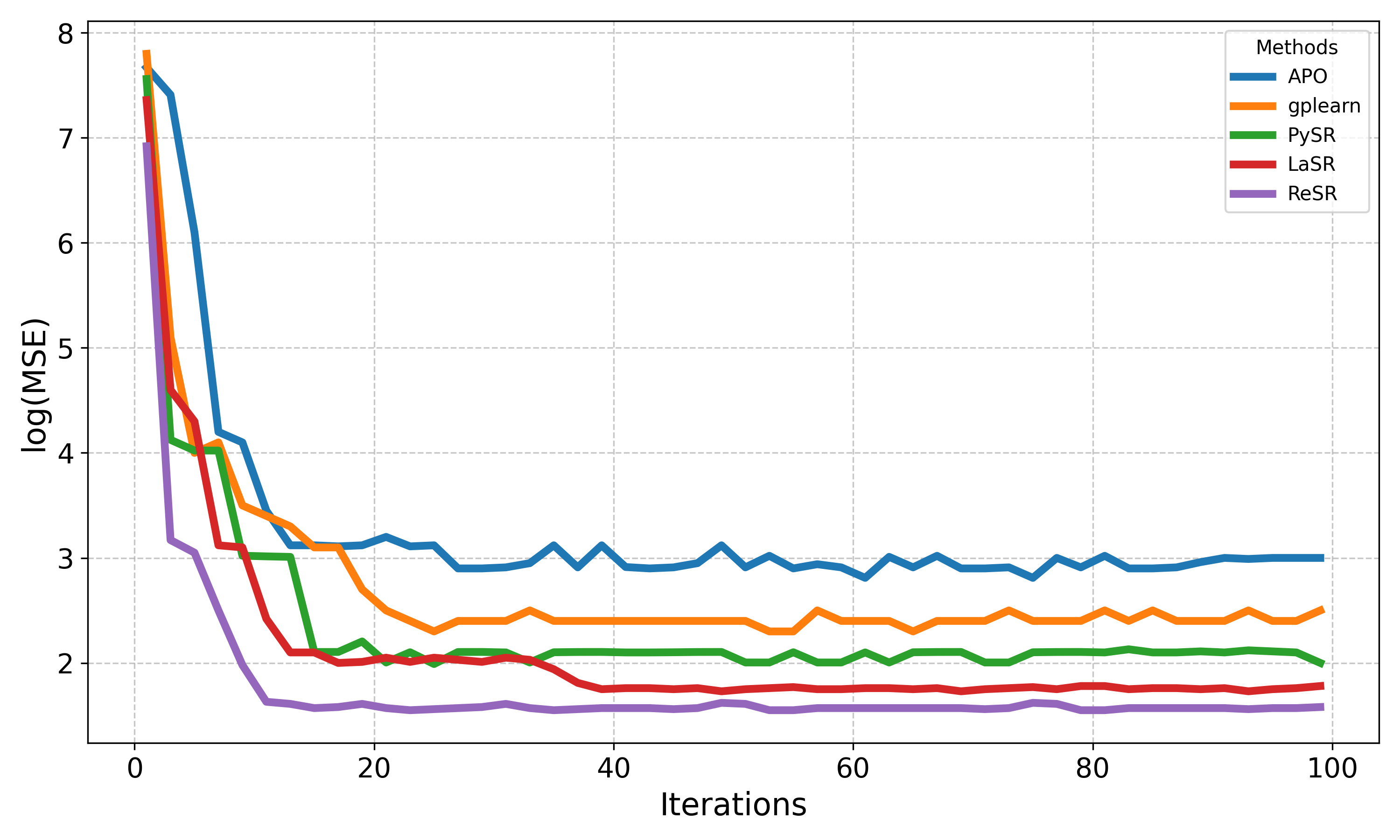}
        \caption{Validation error (MSE) over iterations.}
        \label{fig:SR_convergence_test}
    \end{subfigure}
    
    \caption{Convergence comparison between our method and baselines. We report averaged MSE across physical systems.}
    \label{fig:SR_convergence}
\end{figure}

Figure~\ref{fig:SR_convergence} compares convergence behavior across methods during training and validation. \ourmethod achieves faster convergence and lower MSE earlier in the search process, confirming that retrieval-based initialization offers a strong starting point. While baselines like PySR and LaSR eventually reduce training error, they exhibit noticeably higher validation MSE, suggesting potential overfitting and weaker generalization to future motion.

\begin{figure}[ht]
    \centering
    \begin{subfigure}[a]{\columnwidth}
        \centering
        \includegraphics[width=\columnwidth]{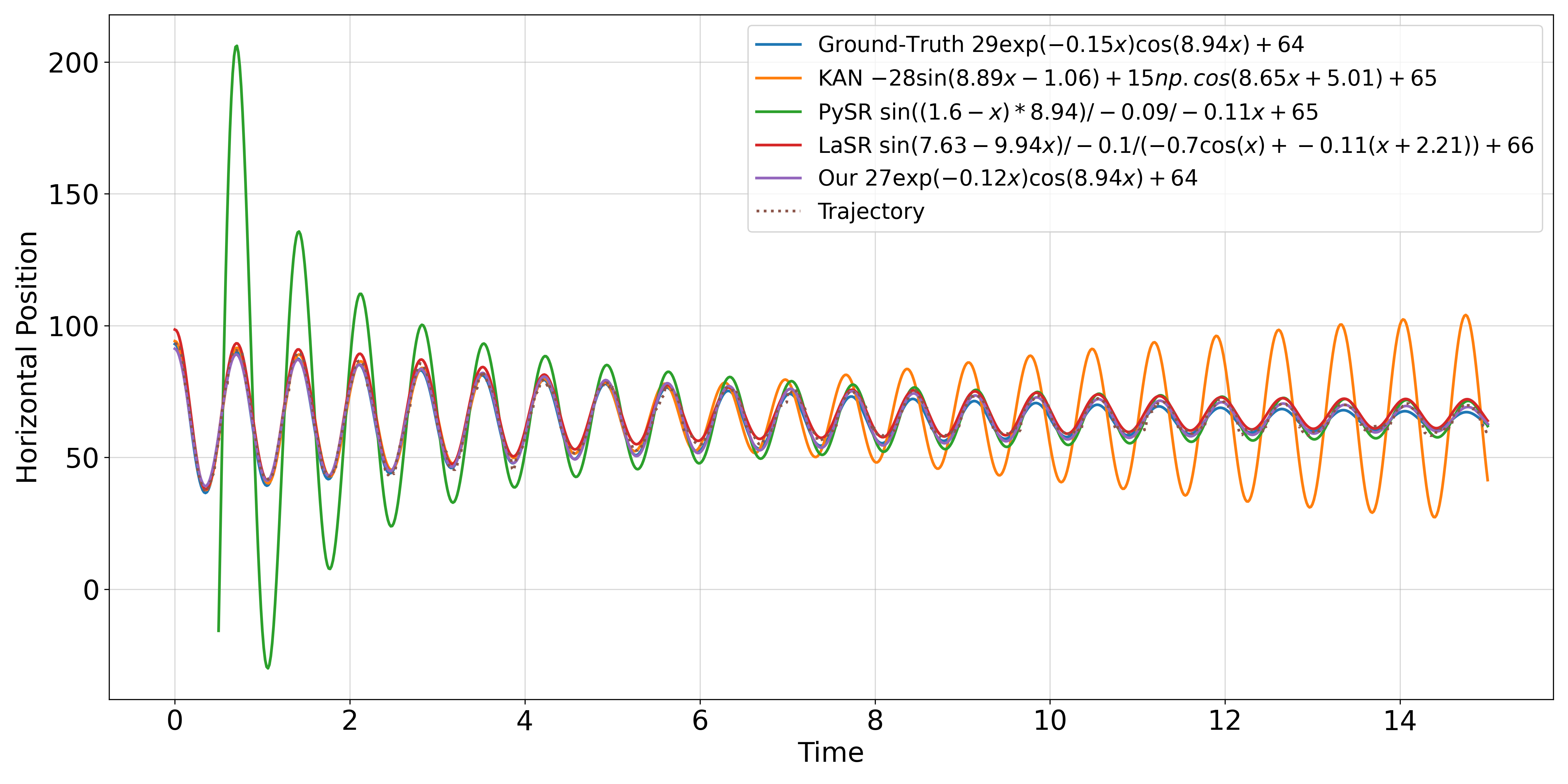}
        \caption{Damped spring-mass system.}
        \label{fig:case_study_spring_mass}
    \end{subfigure}
    
    \vspace{0.1cm}  
    
    \begin{subfigure}[b]{\columnwidth}
        \centering
        \includegraphics[width=\columnwidth]{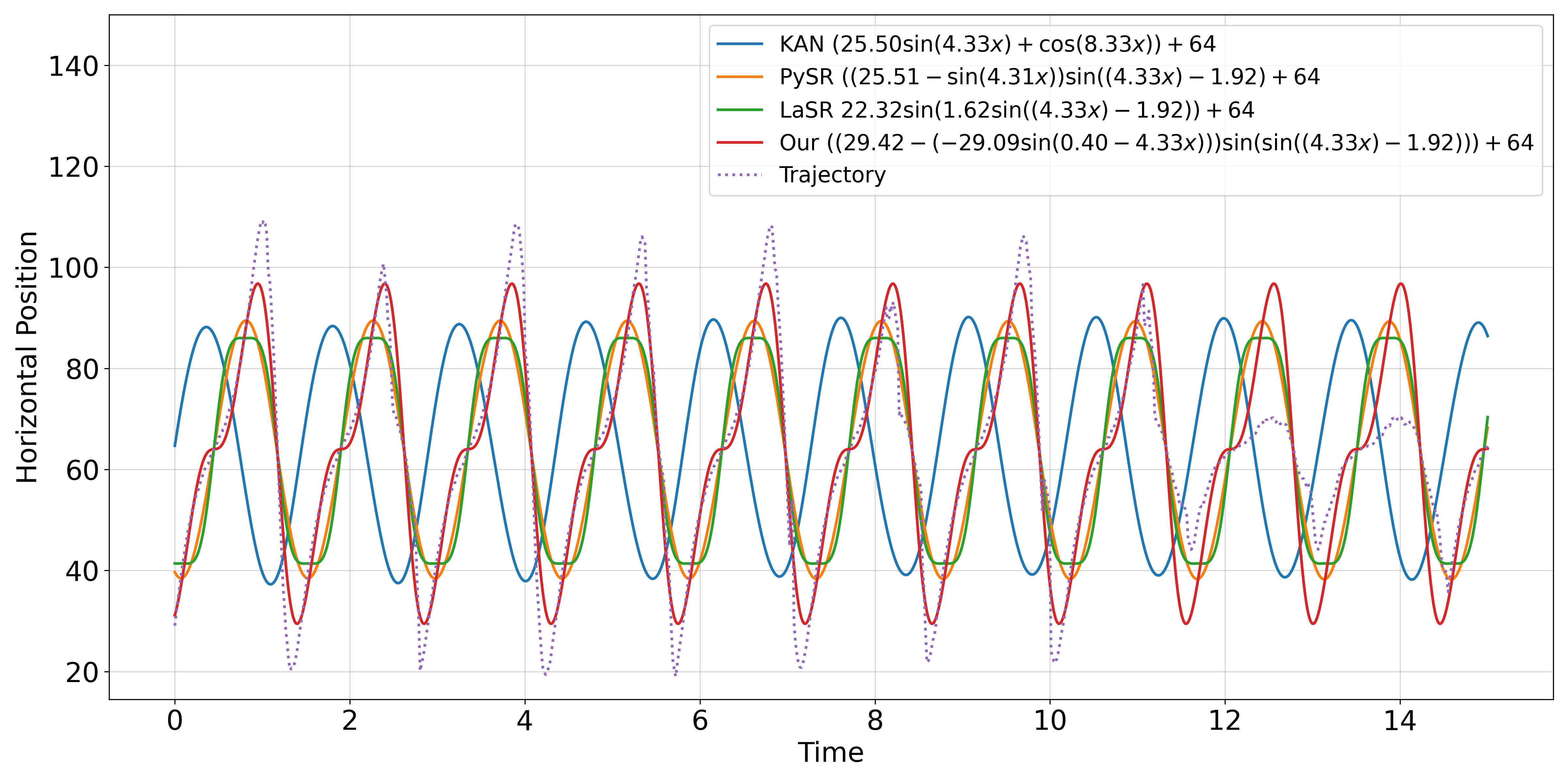}
        \caption{Single pendulum system.}
        \label{fig:case_study_single_pendulum}
    \end{subfigure}
    
    \caption{Comparison of observed trajectories with ground-truth and predicted equations.}
    \label{fig:case_study}
\end{figure}

For qualitative evaluation, Figure~\ref{fig:case_study} presents case studies for damped spring-mass and single pendulum systems. In both cases, \ourmethod produces trajectories that closely follow the observed trajectories. Baselines often yield distorted or phase-shifted trajectories. The performance gap is especially prominent in the single pendulum case (Figure~\ref{fig:case_study_single_pendulum}), where baselines struggle to capture the nonlinearity of single pendulum dynamics, while \ourmethod more accurately recovers the underlying motion dynamics. These results highlight the effectiveness of retrieval-based pre-training for symbolic regression, improving both accuracy and efficiency in discovering interpretable physical equations from observational data.

\subsection{Evaluation of Video Generation}
\paragraph{Datasets}
We evaluate motion forecasting for video generation using the same set of physical systems described in Section~\ref{sec:eval_SR}. Each video generation model takes an initial image (serving as the first frame) and, optionally, a predicted trajectory. Initial images are sourced from both synthetic and real-world domains. Synthetic images are rendered using physics simulators \cite{huang2024automateddiscoverycontinuousdynamics, ohana2024the} and include systems such as spring mass, damped spring mass, two-body, projectile, and fluid motion. Real initial images are extracted from videos of real-world single and double pendulum systems. For each system, we extract ten initial images, each corresponding to the first frame of the final 10\% (test segment) of its video. This setup aligns with our symbolic regression evaluation protocol. To generate future trajectories, we apply the equations learned via \ourmethod or other symbolic regression methods on the training portion (first 80\%) of the trajectory and forecast motion into the test segment.

\paragraph{Evaluation of Generated Videos}
We evaluate generated videos along two axes: visual quality and physics alignment.
For visual quality, we use the Fréchet Video Distance (\textit{FVD}) and Fréchet inception distance (\textit{FID}) \cite{unterthiner2019accurategenerativemodelsvideo, NIPS2017_8a1d6947} to measure the difference between generated and ground-truth videos. Additionally, we employ DINO \cite{9709990} to assess \textit{subject consistency} between the initial image and the generated video. 
For physics alignment, we use AMT \cite{li2023amt} to quantify \textit{motion smoothness}, reflecting the temporal coherence of object motion. We also use \textit{TrajErr} \cite{zhang2024toratrajectoryorienteddiffusiontransformer} to measure the deviation between the input trajectory and the actual object trajectory (extracted by CoTracker) in the generated video. 

We conduct pairwise human comparisons across models for both visual quality and physics alignment. For each model pair, three graduate-level annotators—unaffiliated with model development—independently judged the better video. Each comparison is anonymized and randomized. Evaluations were conducted on ten videos per system across all physical systems. Annotators are provided with system descriptions to aid in assessing physics correctness. Due to budget limitations, human evaluation is restricted to top-performing models based on automatic metrics: CogVideoX1.5 (best trajectory-free baseline), Tora (best open-source trajectory-guided model), Kling (overall best trajectory-guided model), and a physics simulator (for synthetic image only).

\begin{table}[ht]
\centering
\resizebox{\linewidth}{!}{%
\begin{tabular}{@{}llccccc@{}}
\toprule
                                 &                                   & \multicolumn{3}{c|}{\textbf{Visual Quality}}                                                                                                              & \multicolumn{2}{c}{\textbf{Physics Alignment}}                                          \\
\multirow{-2}{*}{\textbf{Class}} & \multirow{-2}{*}{\textbf{Models}} & FVD(↓)                                     & FID(↓)                                   & \multicolumn{1}{c|}{SubC(↑)}                                      & Smooth(↑)                                    & TraErr(↓)                                \\ \midrule
\multicolumn{7}{c}{\textbf{Real Initial Frame}}                                                                                                                                                                                                                                                                            \\ \midrule
                                 & SVD                               & $2521_{ 213}$                         & $396_{ 47}$                         & \multicolumn{1}{c|}{$75.68_{ 2.45}$}                         & $91.52_{ 2.31}$                         & $632_{ 134}$                        \\
                                 & CogVideoX                         & $2203_{ 215}$                         & $356_{ 24}$                         & \multicolumn{1}{c|}{$79.11_{ 1.51}$}                         & $92.40_{ 2.34}$                         & $512_{ 84}$                         \\
                                 & Cosmos                            & $2453_{ 295}$                         & $381_{ 56}$                         & \multicolumn{1}{c|}{$75.48_{ 3.42}$}                         & $91.46_{ 4.35}$                         & $552_{ 78}$                         \\
\multirow{-4}{*}{I2V}            & HunyuanV                          & $2382_{ 274}$                         & $387_{ 52}$                         & \multicolumn{1}{c|}{$76.94_{ 2.63}$}                         & $92.35_{ 3.41}$                         & $578_{ 87}$                         \\ \midrule
I2I                              & ID                                & $2607_{ 221}$                         & $423_{ 31}$                         & \multicolumn{1}{c|}{$71.37_{ 4.53}$}                         & $86.38_{ 4.31}$                         & $563_{ 46}$                         \\ \midrule
                                 & DragAny                           & $1729_{ 133}$                         & $277_{ 35}$                         & \multicolumn{1}{c|}{$78.69_{ 2.54}$}                         & $92.38_{ 2.10}$                         & $489_{ 39}$                         \\
                                 & MotionCtrl                        & $1793_{ 148}$                         & $266_{ 31}$                         & \multicolumn{1}{c|}{$76.32_{ 1.52}$}                         & $91.44_{ 3.08}$                         & $484_{ 31}$                         \\
                                 & SG-I2V                            & $1778_{ 145}$                         & $254_{ 31}$                         & \multicolumn{1}{c|}{$77.10_{ 2.94}$}                         & $92.39_{ 3.84}$                         & $455_{ 40}$                         \\
                                 & Tora                              & $1674_{ 133}$                         & $235_{ 34}$                         & \multicolumn{1}{c|}{$80.28_{ 1.42}$}                         & $95.30_{ 2.28}$                         & $431_{ 37}$                         \\
                                 & Kling                             & \cellcolor[HTML]{FFFFC7}$1064_{ 125}$ & \cellcolor[HTML]{FFFFC7}$194_{ 24}$ & \multicolumn{1}{c|}{\cellcolor[HTML]{FFFFC7}$84.63_{ 1.84}$} & \cellcolor[HTML]{FFFFC7}$97.53_{ 2.41}$ & \cellcolor[HTML]{FFFFC7}$404_{ 40}$ \\
                                 & Kling-LaSR                        & $1225_{ 137}$                         & $211_{ 32}$                         & \multicolumn{1}{c|}{$84.14_{ 1.43}$}                         & $96.98_{ 2.10}$                         & $451_{ 43}$                         \\
                                 & Kling-Manual                      & $1329_{ 129}$                         & $218_{ 28}$                         & \multicolumn{1}{c|}{\cellcolor[HTML]{96FFFB}$85.27_{ 2.11}$} & $97.17_{ 2.33}$                         & $477_{ 36}$                         \\
\multirow{-8}{*}{TrajI2V}        & Kling-CoTracker                   & \cellcolor[HTML]{96FFFB}$1022_{ 132}$ & \cellcolor[HTML]{96FFFB}$186_{ 31}$ & \multicolumn{1}{c|}{$84.46_{ 1.31}$}                         & \cellcolor[HTML]{96FFFB}$97.88_{ 2.18}$ & \cellcolor[HTML]{96FFFB}$397_{ 34}$ \\ \midrule
\multicolumn{7}{c}{\textbf{Synthetic Initial Frame}}                                                                                                                                                                                                                                                                       \\ \midrule
                                 & SVD                               & $1947_{ 294}$                         & $415_{ 45}$                         & \multicolumn{1}{c|}{$79.07_{ 2.48}$}                         & $95.24_{ 2.41}$                         & $624_{ 121}$                        \\
                                 & CogVideoX                         & $1547_{ 154}$                         & $342_{ 31}$                         & \multicolumn{1}{c|}{$85.75_{ 2.49}$}                         & $97.38_{ 1.49}$                         & $589_{ 93}$                         \\
                                 & Cosmos                            & $1870_{ 215}$                         & $388_{ 41}$                         & \multicolumn{1}{c|}{$81.39_{ 4.31}$}                         & $96.39_{ 1.40}$                         & $658_{ 74}$                         \\
\multirow{-4}{*}{I2V}            & HunyuanV                          & $1753_{ 186}$                         & $363_{ 39}$                         & \multicolumn{1}{c|}{$83.31_{ 3.19}$}                         & $96.38_{ 1.57}$                         & $562_{ 63}$                         \\ \midrule
I2I                              & ID                                & $1994_{ 127}$                         & $348_{ 43}$                         & \multicolumn{1}{c|}{$77.44_{ 3.13}$}                         & $89.39_{ 5.95}$                         & $567_{ 81}$                         \\ \midrule
                                 & DragAny                           & $853_{ 142}$                          & $248_{ 35}$                         & \multicolumn{1}{c|}{$79.69_{ 2.14}$}                         & $97.84_{ 1.01}$                         & $374_{ 32}$                         \\
                                 & MotionCtrl                        & $847_{ 153}$                          & $244_{ 27}$                         & \multicolumn{1}{c|}{$80.52_{ 3.31}$}                         & $97.43_{ 1.45}$                         & $378_{ 45}$                         \\
                                 & SG-I2V                            & $791_{ 138}$                          & $235_{ 24}$                         & \multicolumn{1}{c|}{$81.48_{ 2.93}$}                         & $96.53_{ 1.46}$                         & $384_{ 33}$                         \\
                                 & Tora                              & $728_{ 121}$                          & $193_{ 28}$                         & \multicolumn{1}{c|}{$85.21_{ 2.49}$}                         & $98.43_{ 0.55}$                         & $354_{ 29}$                         \\
                                 & Kling                             & \cellcolor[HTML]{FFFFC7}$641_{ 103}$  & \cellcolor[HTML]{FFFFC7}$135_{ 27}$ & \multicolumn{1}{c|}{\cellcolor[HTML]{FFFFC7}$89.49_{ 3.42}$} & \cellcolor[HTML]{96FFFB}$98.93_{ 0.41}$ & \cellcolor[HTML]{FFFFC7}$325_{ 30}$ \\
                                 & Kling-LaSR                        & $710_{ 113}$                          & $151_{ 31}$                         & \cellcolor[HTML]{96FFFB}$89.63_{ 2.28}$                      & $98.24_{ 0.53}$                         & $341_{ 31}$                         \\
                                 & Kling-Manual                      & $667_{ 127}$                          & $164_{ 25}$                         & \multicolumn{1}{c|}{$89.17_{ 2.77}$}                         & \cellcolor[HTML]{FFFFC7}$98.77_{ 0.38}$ & $357_{ 27}$                         \\
\multirow{-8}{*}{TrajI2V}        & Kling-CoTracker                   & \cellcolor[HTML]{96FFFB}$633_{ 105}$  & \cellcolor[HTML]{96FFFB}$132_{ 24}$ & \multicolumn{1}{c|}{$88.43_{ 3.10}$}                         & $98.36_{ 0.47}$                         & \cellcolor[HTML]{96FFFB}$320_{ 34}$ \\ \cmidrule(l){2-7} 
\rowcolor[HTML]{EFEFEF} 
Rule                             & Simulator                         & $67_{ 12}$                            & $16_{ 3}$                           & \multicolumn{1}{c|}{\cellcolor[HTML]{EFEFEF}$94.25_{ 2.83}$} & $99.52_{ 0.01}$                         & $48_{ 14}$                          \\ \bottomrule
\end{tabular}%
}
\caption{Evaluation of video forecasting. All reported metrics are averaged across physical systems. SubC and Smooth represent subject consistency and motion smoothness, reported by \%. The simulator serves as a reference for upper-bound performance on synthetic initial frame setting. The top two results are highlighted as \colorbox[HTML]{38FFF8}{best} and \colorbox[HTML]{FFFFC7}{second}, respectively.}
\label{tab:eval_video}
\vspace{-15pt}
\end{table}

\paragraph{Video Generation Models}
We conduct experiments on trajectory-free and trajectory-guided I2V models. For trajectory-free baselines, we consider state-of-the-art I2V models, including SVD \cite{blattmann2023stable}, CogVideoX1.5 \cite{yang2025cogvideox}, Cosmos \cite{nvidia2025cosmosworldfoundationmodel}, and HunyuanVideo \cite{kong2025hunyuanvideosystematicframeworklarge}, which generate future frames directly from a single initial image and, optionally, text prompt (\eg CogVideoX1.5, Cosmos and HunyuanVideo). We include ID \cite{chen2022automated}, an encoder-decoder model that generates videos frame-by-frame without trajectory guidance. For trajectory-guided I2V models, we use DragAnything \cite{10.1007/978-3-031-72670-5_19}, MotionCtrl \cite{10.1145/3641519.3657518}, SG-I2V \cite{namekata2025sgiv}, Tora \cite{zhang2024toratrajectoryorienteddiffusiontransformer} and Kling (a commercial model) \cite{klingai}. These models are conditioned on the initial image and the predicted trajectories generated by learned equations. Some models, such as Tora and Kling, additionally accept a text prompt for further guidance. For synthetic data, we also use a physics simulator \cite{huang2024automateddiscoverycontinuousdynamics} to generate future videos. However, since it is not scalable to real-world scenarios, we include it only for comparison in synthetic settings, serving as a reference for upper-bound performance.

\paragraph{Implementation Details}
We resize initial images to match each model's input resolution. The video length is fixed at 5 seconds, with frames per second (FPS) set per model requirements. For models requiring text prompts, we use either official prompt guidelines or generate prompts using GPT-4o \cite{yang2025cogvideox}. Trajectories are normalized and scaled to match each model's spatial resolution and sampled uniformly at 2 points per second. All models are run on a machine with an NVIDIA 80G A100 GPU with publicly released checkpoints or API without additional fine-tuning.

\paragraph{Results and Analysis}

Table~\ref{tab:eval_video} presents automatic evaluation results across all models and physical systems. Models guided by predicted trajectories consistently outperform trajectory-free baselines in both visual quality and physics alignment. Among trajectory-guided I2V models, Kling achieves the strongest overall performance. However, a notable performance gap remains compared to the physics simulator, which we treat as an upper bound. For instance, in the synthetic setting, Kling achieves an FVD of 641 and a TraErr of 325, while the simulator reaches significantly lower values of 67 and 48, respectively. Real initial frame settings have lower performance than synthetic settings, likely due to background noises and systems complexity, such as the double pendulum. These results highlight the limitations of current diffusion-based video generation models in accurately capturing physical dynamics—even when guided by physics-aligned trajectories.

We perform an ablation study comparing Kling guided by different trajectory sources. Trajectories predicted by \ourmethod outperform both manually drawn trajectories and those derived from LaSR, which is the best symbolic regression baseline, demonstrating that our approach more faithfully captures the underlying physical dynamics. Furthermore, Kling guided by \ourmethod achieves performance comparable to Kling using ground-truth future trajectories (extracted by CoTracker), highlighting the small gap between trajectories predicted by \ourmethod and ground-truth future trajectories.

\begin{table}[ht]
\centering
\resizebox{\columnwidth}{!}{%
\begin{tabular}{lcc}
\hline
\textbf{Comparison}    & \textbf{Visual Quality} & \textbf{Physics Alignment} \\ \hline
\multicolumn{3}{c}{\textbf{Real Initial Frame}}                               \\ \hline
\textbf{Tora} vs. CogVideoX     & 61\%                    & 84\%                       \\
\textbf{Kling} vs. Tora         & 67\%                    & 71\%                       \\
\textbf{Kling} vs. Kling-manual & 57\%                    & 63\%                       \\ \hline
\multicolumn{3}{c}{\textbf{Synthetic Initial Frame}}                          \\ \hline
\textbf{Tora} vs. CogVideoX     & 78\%                    & 77\%                       \\
\textbf{Kling} vs. Tora         & 66\%                    & 61\%                       \\
\textbf{Kling} vs. Kling-manual & 59\%                    & 64\%                       \\
\textbf{Simulator} vs. Kling    & 94\%                    & 97\%                       \\ \hline
\end{tabular}%
}
\caption{Human evaluation via pairwise comparisons. In each row, the \textbf{bolded} model indicates the winner, and the following cell reports its win rate for each criterion.}
\label{tab:human_eval}
\vspace{-15pt}
\end{table}

Human evaluation results in Table~\ref{tab:human_eval} support these findings. Across pairwise comparisons, annotators consistently preferred trajectory-guided models over trajectory-free baseline on both physical alignment and visual quality. Notably, Kling with \ourmethod-guided trajectories was preferred over its manually guided counterpart, confirming that learned equations offer more accurate and reliable motion control. In the synthetic setting, the physics simulator was consistently rated as the most physically accurate, highlighting the limitations of data-driven models. We attribute this to a fundamental difference: current video generation models are trained to capture statistical correlations in large-scale datasets, but lack explicit modeling of physical causality. In contrast, physics simulators generate motion directly from governing equations, ensuring high physical fidelity. However, simulators have their own limitations. They are not scalable across diverse scenarios and tend to lack realism when applied to real-world scenarios. This highlights the value of our method, which seeks to combine the interpretability and physical grounding of governing equations with the flexibility and realism of data-driven generative models.

\begin{figure}[ht]
    \centering
    \begin{subfigure}[a]{\columnwidth}
        \centering
        \includegraphics[width=0.99\columnwidth]{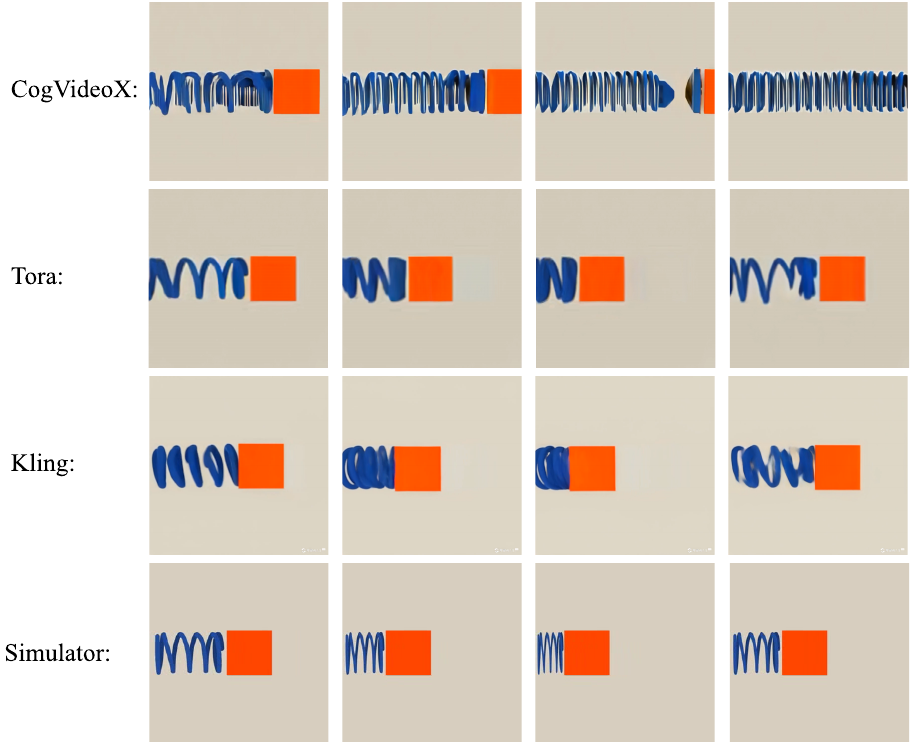}
        \caption{Damped spring-mass system (Synthetic).}
        \label{fig:video_case_study_spring_mass}
    \end{subfigure}
    
    \vspace{0.1cm}  
    
    \begin{subfigure}[b]{\columnwidth}
        \centering
        \includegraphics[width=\columnwidth]{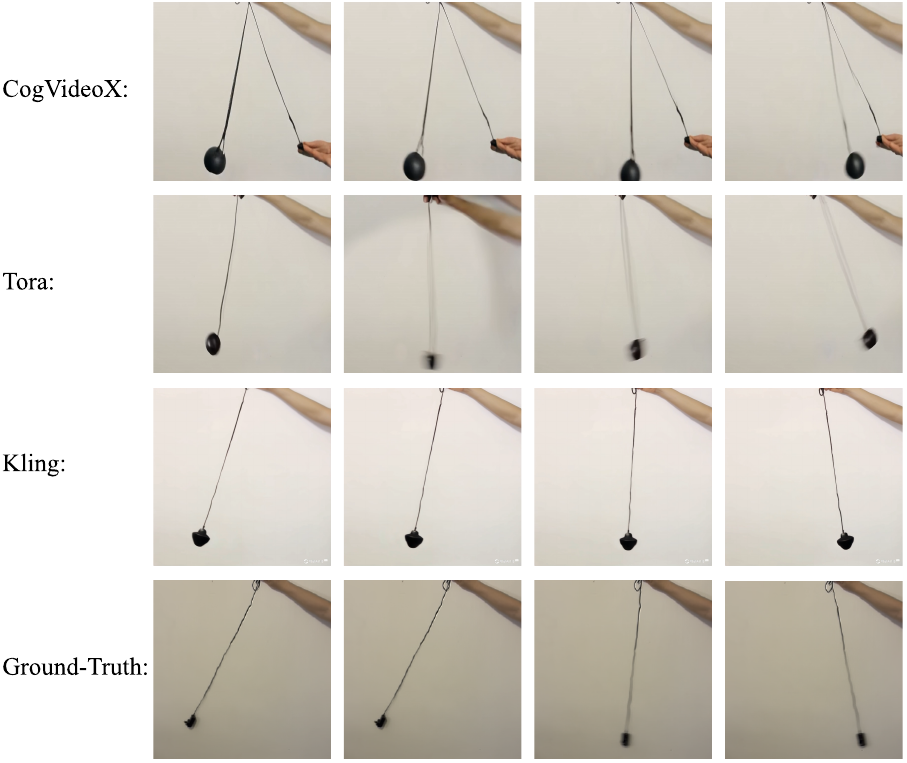}
        \caption{Single pendulum system (Real).}
        \label{fig:video_case_study_single_pendulum}
    \end{subfigure}
    
    \caption{Qualitative comparisons across models, showcasing frames from different models at the same time step for visual evaluation. }
    \label{fig:video_case_study}
\vspace{-15pt}
\end{figure}

Figure~\ref{fig:video_case_study} illustrates qualitative comparisons. Trajectory-guided models exhibit improved global motion consistency, while trajectory-free models (e.g., CogVideoX) often produce erratic or implausible dynamics. Even the strongest model, Kling, fails to capture fine-grained physical details such as spring deformation, suggesting that while trajectory conditioning improves high-level motion, current models still lack the physical inductive biases needed for fine-grained dynamic synthesis.



\section{Related Work}

\paragraph{Learning Physics from Video}
Several works aim to extract physical laws or parameters of dynamic systems directly from video, using symbolic regression or ODE-based methods \cite{chari2019visualphysicsdiscoveringphysical, luan2021uncoveringclosedformgoverningequations, Tetriyani_Jihad_Karlina, garcia2024learningphysicsvideounsupervised}. However, many of these approaches impose strong constraints on the equation structure, such as assuming linearity, or focus solely on estimating parameters of pre-defined models. Other methods use autoencoders to encode video sequences into low-dimensional latent vectors and attempt to learn system dynamics in that space \cite{huang2024automateddiscoverycontinuousdynamics}. These latent variables often lack physical interpretability, and the resulting dynamics are not expressed as symbolic equations. In contrast, our approach employs symbolic regression to directly learn explicit symbolic equations, capturing physically meaningful variables that map time to object positions, thus ensuring interpretability and physical alignment.

\paragraph{Physics-aligned Video Generation}
To ensure physical realism in video generation, one common approach is to use physics simulators, where dynamics are modeled via hard-coded rules and differential equations \cite{millington2007game, 6386109, bonnet2022airfrans, kohl2024turbulent, ohana2024the}. While highly accurate, these simulators are typically limited to specific domains, require hand-crafted scenario design, and often demand significant computational resources and time consumption. On the other hand, recent video generation models based on diffusion or autoregressive architectures \cite{blattmann2023stable, wang2024emu3nexttokenpredictionneed, yang2025cogvideox, nvidia2025cosmosworldfoundationmodel, kong2025hunyuanvideosystematicframeworklarge} can synthesize diverse scenes from image or text prompts but often lack physical consistency. These models are trained on large-scale datasets without the ability to interact with the real world and thus cannot observe the causal effects by interventions, leading to unrealistic object motion \cite{motamed2025generativevideomodelsunderstand}. Our method bridges this gap by integrating symbolic equation learning into the video generation pipeline. Rather than relying solely on data-driven models, we first use object-tracking tools to extract motion trajectories, then apply symbolic regression to learn equations of motion. These equations are used to predict future trajectories, which serve as guidance signals for trajectory-guided video generation, resulting in videos that better align with real-world physics.

\section{Conclusion}
We introduce a novel neuro-symbolic, physics-grounded, inference-only framework for motion forecasting in trajectory-guided video generation, which employs a retrieval-enhanced SR algorithm, \ourmethod, for equation discovery. Experimental results demonstrate that our approach can reliably generate future motion trajectories closely matching equations derived from Classical Mechanics, while also highlighting the strengths and limitations of the SOTA I2V models on this task. Overall, our work illustrates the potential of integrating interpretable equation discovery with generative models and paves the way for future applications in scientific discovery and world simulations for robotics. 

\bibliographystyle{ACM-Reference-Format}
\bibliography{custom}


\end{document}